\documentclass[journal,twoside,web]{ieeecolor}
\usepackage{tmi}

\usepackage{etoolbox}
\makeatletter
\@ifundefined{color@begingroup}%
  {\let\color@begingroup\relax
   \let\color@endgroup\relax}{}%
\def\fix@ieeecolor@hbox#1{%
  \hbox{\color@begingroup#1\color@endgroup}}
\patchcmd\@makecaption{\hbox}{\fix@ieeecolor@hbox}{}{\FAILED}
\patchcmd\@makecaption{\hbox}{\fix@ieeecolor@hbox}{}{\FAILED}
\usepackage[table]{xcolor}

\usepackage{cite}
\usepackage{amsmath,amssymb,amsfonts}
\usepackage{bm}
\usepackage{relsize}
\usepackage{hyperref}
\usepackage{graphicx}
\usepackage{textcomp}
\renewcommand{\textapprox}{\raisebox{0.5ex}{\texttildelow}}
\usepackage{algorithm}
\usepackage{algpseudocode}
\usepackage{tabularx,booktabs}

\usepackage{multirow}
\usepackage{comment}
\let\labelindent\relax
\usepackage[inline]{enumitem}
\DeclareMathOperator*{\argmin}{argmin}

\newcommand{\cinema}{\emph{CINeMA}}
\newcommand{\circumeq}{\mathrel{\widehat{=}}}
\usepackage{pifont}
\newcommand{\cmark}{\ding{51}}%
\newcommand{\xmark}{\ding{55}}%
\newcolumntype{C}{>{\centering\arraybackslash}X} 
\def\BibTeX{{\rm B\kern-.05em{\sc i\kern-.025em b}\kern-.08em
    T\kern-.1667em\lower.7ex\hbox{E}\kern-.125emX}}
\begin{document}
\title{
CINeMA: Conditional Implicit Neural Multi-Modal Atlas for a Spatio-Temporal Representation of the Perinatal Brain}

\author{Maik Dannecker, Vasiliki Sideri-Lampretsa, Sophie Starck, Angeline Mihailov, Mathieu Milh, Nadine Girard, Guillaume Auzias, and Daniel Rueckert, \IEEEmembership{Fellow, IEEE}
\thanks{(Corresponding author: Maik Dannecker. These authors jointly supervised this work: Guillaume Auzias, Daniel Rueckert)}
\thanks{This work was supported by the ERC (Deep4MI-884622), the ERA-NET NEURON Cofund (MULTI-FACT-8810003808), and the French State as part of France 2030 investment plan, part of Aix-Marseille University Excellence Initiative A*MIDEX (AMX-19-IET-002). Maik Dannecker, Vasiliki Sideri-Lampretsa, Sophie Starck, and Daniel Rueckert are with the School of Computation, Information and Technology, and the School of Medicine and Health, Technical University of Munich, Germany (e-mail: \{m.dannecker, vasiliki.sideri-lampretsa, sophie.starck, daniel.rueckert\}@tum.de). Daniel Rueckert is also with the Department of Computing, Imperial College London, United Kingdom. Angeline Mihailov, Nadine Girard, Guillaume Auzias are with the Institut de Neurosciences de la Timone, UMR 7289, CNRS, Aix-Marseille Université, 13005, Marseille, France (e-mail:{angeline.mihailov, guillaume.auzias}@univ-amu.fr). Nadine Girard is also with the Aix-Marseille Univ, APHM, Service de Neuroradiologie Diagnostique et Interventionnelle, Hôpital de la Timone 2, 13005, Marseille, France (e-mail:Nadine.GIRARD@ap-hm.fr). Mathieu Milh is with the Aix-Marseille Univ, APHM, service de neurologie pédiatrique, Hôpital de la Timone, 13005, Marseille, France (e-mail:Mathieu.milh@ap-hm.fr ).}}

\maketitle

\begin{abstract}
Magnetic resonance imaging of fetal and neonatal brains reveals rapid neurodevelopment marked by substantial anatomical changes unfolding within days. Studying this critical stage of the developing human brain, therefore, requires accurate brain models\textemdash{}referred to as atlases\textemdash{}of high spatial and temporal resolution. To meet these demands, established traditional atlases and recently proposed deep learning-based methods rely on large and comprehensive datasets. This poses a major challenge for studying brains in the presence of pathologies for which data remains scarce.  We address this limitation with \cinema~(Conditional Implicit Neural Multi-Modal Atlas), a novel framework for creating high-resolution, spatio-temporal, multimodal brain atlases, suitable for low-data settings. Unlike established methods, \cinema~operates in latent space, avoiding compute-intensive image registration and reducing atlas construction times from days to minutes. Furthermore, it enables flexible conditioning on anatomical features including GA, birth age, and pathologies like ventriculomegaly (VM) and agenesis of the corpus callosum (ACC). \cinema~supports downstream tasks such as tissue segmentation and age prediction whereas its generative properties enable synthetic data creation and anatomically informed data augmentation. Surpassing state-of-the-art methods in accuracy, efficiency, and versatility, \cinema~represents a powerful tool for advancing brain research. We release the code and atlases at \href{https://github.com/m-dannecker/CINeMA}{https://github.com/m-dannecker/CINeMA}.

\end{abstract}

\begin{IEEEkeywords}
conditional atlas, perinatal brain imaging, implicit neural representation, magnetic resonance imaging, segmentation 
\end{IEEEkeywords}

\section{Introduction} \label{sec:introduction}
\IEEEPARstart {T}{he} spatio-temporal analysis of perinatal brain structures from magnetic resonance imaging (MRI) is crucial for understanding normal and abnormal early brain development. Brain atlases serve as important tool in this analysis, offering prior anatomical knowledge to enhance tasks such as segmentation, shape analysis, and the characterization of both typical and atypical neurodevelopmental patterns \cite{ashburner2007fast, fischl2012freesurfer, Avants2009, UusBounty2023, Ciceri2024, Serag2012, Gholipour2017}. The rapid neurodevelopment of perinatal brains, including cortical folding and growth, highlights the need for detailed, high-resolution atlases capturing anatomical changes in this critical stage. Publicly available datasets, such as FeTA \cite{FeTA2021} and dHCP \cite{price2019dhcpACQ, Edwards2022}, provide hundreds of fetal and neonatal brain scans, facilitating the construction of atlases for studying early \emph{healthy} brain development.

Despite these advancements, comprehensive datasets of perinatal brains with abnormal development remain scarce. This challenges traditional atlas construction methods, which rely on large, age-balanced datasets to generate unbiased spatio-temporal brain atlases. Moreover, these methods lack conditional properties, i.e., the ability to adapt the atlas to additional factors such as sex or anatomical abnormalities like ventriculomegaly (VM), proving them inadequate to characterize pathological growth patterns. Deep learning-based approaches for atlas construction have emerged as one alternative, leveraging spatial transformers and generative adversarial networks (GANs) \cite{dey2021generative, Dalca2019, Sinclair2022}. 
These methods advertise conditioning on features like GA, sex or pathology; however, their application to pathological brains or specific abnormalities remains limited. Additionally, their compute-intensive and often unstable training processes\textemdash{}GANs, for example, may suffer from oscillations or mode collapse\textemdash{}combined with their reliance on large training datasets, pose significant limitations.

\subsection{Contribution}
We introduce \cinema~(\textbf{C}onditional \textbf{I}mplicit \textbf{Ne}ural \textbf{M}ultimodal \textbf{A}tlas), an extended framework of CINA (Conditional Implicit Neural Atlas)\cite{Dannecker2024CINA}, for creating high-resolution, spatio-temporal, and multimodal brain atlases. By leveraging implicit neural representations (INRs), \cinema~offers a continuous, resolution-agnostic atlas that captures typical and pathological anatomy with minimal data. Unlike established methods that rely on computationally intensive deformation fields to account for subject variability, \cinema~encodes individual variability directly into subject-specific latent codes via an auto-decoder setup. This eliminates the need for explicit deformable registration, reducing atlas construction time from days to minutes, while constructing atlases of superior representation accuracy compared to established baselines. Furthermore, \cinema~enables flexible conditioning on anatomical features, including GA and abnormalities such as VM and agenesis of the corpus callosum (ACC), while supporting downstream tasks like tissue segmentation and age prediction. Its generative properties enable synthetic data creation and anatomically informed data augmentation, facilitated by a smooth and compact latent space. 

Building upon the prior work of CINA \cite{Dannecker2024CINA}, this work's extensions include multimodal integration, architectural enhancements for improved atlas quality and accuracy, and more extensive experiments involving two additional datasets, including a new domain focused on neonatal brains.

To summarize, \cinema~provides:
\begin{itemize}
    \item A high-resolution, spatio-temporal, multi-modal atlas generated in minutes rather than days, using images of heterogeneous resolution and modality.
    \item A novel end-to-end framework, bypassing affine and deformable registration while capturing anatomical variability in a dedicated latent space.
    \item Robust performance in low data regimes, succeeding where established methods fail.
    \item Synthetic data creation and anatomically informed data augmentation.
    \item A versatile conditioning framework for modeling typical and atypical neurodevelopment, including abnormal anatomical features like enlarged lateral ventricles or absence of the corpus callosum.
\end{itemize}

\section{Related Work}

\subsection{Traditional Atlas Construction}
The first works on probabilistic anatomical atlases from MRI used affine registration to align images to a reference space defined by a single subject \cite{evans19933d, mazziotta2001, talairach19883}. To mitigate the inherent bias toward the chosen reference subject, group-wise registration techniques were developed, applying iterative estimates and updates of the reference space during atlas construction \cite{bhatia2004group}. Early perinatal brain atlases leveraged these techniques, with notable examples including the fetal atlas by Habas et al. \cite{Habas2010} and the pre-term neonatal atlas by Kuklisova-Murgasova et al. \cite{Kuklisova2011}. These atlases introduced spatio-temporal designs to account for rapidly evolving brain anatomy, creating atlases for predefined GA intervals. For example, the fetal atlas covered 20-25 weeks GA using 20 subjects, while the neonatal atlas spanned 29-44 weeks GA and utilized 142 preterm subjects.

Subsequent advances introduced non-rigid registration to enhance anatomical definition. Serag et al. \cite{Serag2012} employed age-adaptive kernels for weighting subject contributions, producing sharper atlases. Makropoulos et al. \cite{Makropoulus2016} later constructed a neonatal atlas from 420 neonates, using non-rigid free-form deformations \cite{rueckert1999nonrigid} to align images to a common reference space. Gholipour et al. \cite{Gholipour2017} followed by developing a spatio-temporal atlas for 81 healthy fetuses, spanning 21-37 weeks GA. These atlases provide brain tissue segmentation with varying levels of granularity, ranging from broader categorizations such as cortical and subcortical regions \cite{Serag2012}, to more detailed anatomical structures including over hundred regions \cite{Gholipour2017}.

However, traditional atlas construction methods face major challenges. For instance, these methods create static atlases for predefined discrete time points (typically one per week), requiring the entire pipeline to be re-computed for new time points, which is both time-consuming and inefficient\cite{Kuklisova2011, Serag2012, Gholipour2017, Makropoulus2016}.
Furthermore, these approaches require large amounts of data as they lack the ability to leverage information across the dataset, reducing their adaptability and practical value. For example, constructing an atlas for a specific condition, such as a particular GA, demands sufficiently large numbers of subjects to ensure general representations. The issue becomes more pronounced when introducing additional conditions, e.g., volume of the lateral ventricle (LV). Conditions and GA then span a multi-dimensional space of combinations. To generate accurate atlases, traditional approaches therefore require datasets that exponentially grow with the number of conditions. Hence, in the low-data setting of perinatal neuroimaging, even modeling two conditions becomes unfeasible. Consequently, these atlases are mostly constrained to representing healthy anatomy, rendering them inadequate for capturing pathological or atypical growth patterns. These shortcomings underscore the urgent need for data-efficient methodologies that integrate spatial and temporal dimensions and incorporate anatomical conditioning.

\subsection{Deep Learning-Based Atlases}

Image registration has seen major advancements with the adoption of deep learning approaches. Early works like Spatial Transformer Networks (STN) \cite{jaderberg2015spatial} introduced frameworks for supervised and unsupervised end-to-end learning of spatial transformations. Follow-up work such as VoxelMorph \cite{Vxlmrph2019} and DLIR \cite{de2019deep} utilized encoder-decoder architectures to predict displacement fields, automating the registration process and propagating label maps for segmentation tasks. With the success of image registration, deep learning-based atlas construction has emerged as a promising area of research. Combining the concepts of traditional atlas construction with deep learning-based registration tools, inspired numerous new methods that register the training cohort to a common reference space for atlas creation \cite{Ding_2022_CVPR, Dalca2019, Sinclair2022, liu2021cas}. Jointly learning registration and segmentation further leveraged anatomical shape priors to generalize across heterogeneous datasets \cite{Sinclair2022, liu2021cas}. For instance, Dalca et al. \cite{Dalca2019} developed a framework that simultaneously learns conditional templates and deformation fields, improving modeling flexibility. Dey et al. \cite{dey2021generative} expanded the framework using generative adversarial networks (GANs), introducing a discriminator to refine the generated templates. These approaches model the temporal dimension as a continuous variable, offering greater flexibility than traditional discrete methods. Additionally, they can incorporate extra conditions, e.g., LV volume.

Despite these advancements, deep learning methods face challenges. They often require large datasets to achieve robust generalization, demand high computational resources, and may struggle with stability during training, especially GAN-based approaches\cite{dey2021generative}. Moreover, existing conditional methods often lack corresponding brain tissue maps, requiring additional post-processing steps for segmentation \cite{Dalca2019, dey2021generative}. These limitations highlight the ongoing need for improvements in low-data and pathological scenarios.

\subsection{Implicit Neural Representation for Atlas Construction}
Implicit Neural Representations (INRs) offer a new direction for spatio-temporal atlas construction by representing data as continuous coordinate functions via multilayer perceptrons (MLPs) \cite{Mildenhall2021, Siren}. Unlike conventional voxel-based methods, INRs offer a resolution-agnostic framework. Conditioning architectures enhance flexibility by incorporating latent codes, which can be concatenated with input coordinates \cite{Park2019}, linearly projected to modulate MLP layers \cite{Mehta, Functa2022}, or mapped to network weights via hypernetworks \cite{SitzmanConditioned}. Auto-decoder techniques optimize these latent codes iteratively, facilitating efficient modeling of spatio-temporal variations. These properties make INRs particularly suitable for sparse and heterogeneous datasets, addressing gaps left by traditional and deep learning-based methods.
Early applications leverage only parts of these properties. Chen et al. \cite{ChenINRAtlas20222} used INRs to enhance spatial and temporal resolution in fetal brain atlases, while Großbröhmer et al. \cite{Sina2024} modeled subject variability with individualized displacement fields. However, these methods overlook INR conditioning, which is crucial for modeling pathological anatomy and unlocking downstream applications. Dannecker et al. \cite{Dannecker2024CINA} introduced the first Conditional Implicit Neural Atlas (CINA), enabling continuous spatio-temporal atlas construction while incorporating pathologies like VM and supporting tasks like tissue segmentation and age prediction.

With \cinema, we extend this framework by enhancing atlas quality, improving segmentation accuracy, and integrating multimodality. Our expanded experiments introduce two new datasets for neonatal and pathological fetal data, demonstrating \cinema's capability to accurately model preterm neonates and rare fetal abnormalities like ventriculomegaly and ACC.

\section{Methodology}

\begin{figure*}[ht]
    \centerline{\includegraphics[width=\textwidth]{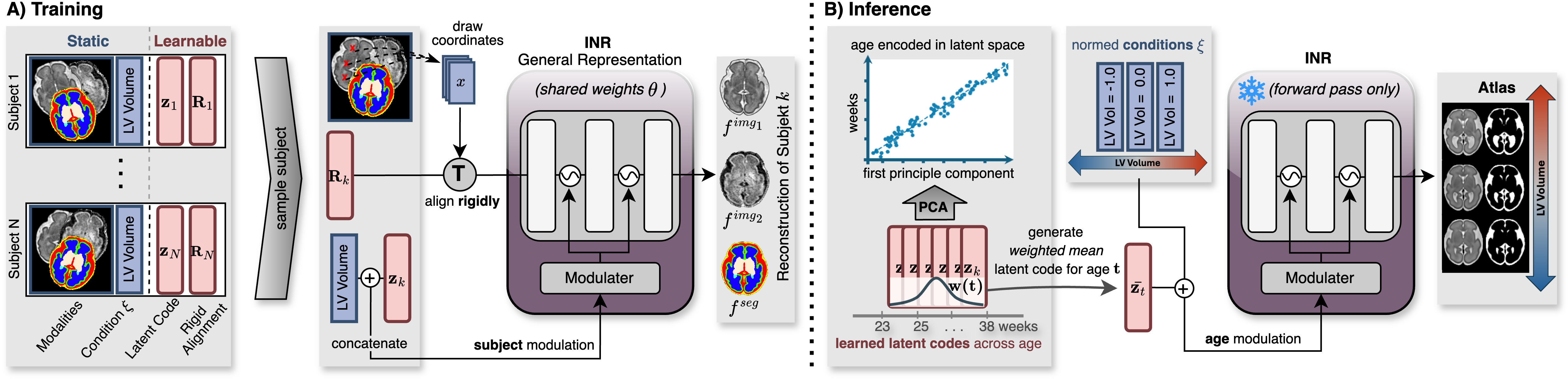}}
   \caption{
        Overview of atlas construction with \cinema.
        \textbf{(i) Training} is conducted on $N$ subjects, each represented by image modalities, tissue segmentation maps, and randomly initialized latent code ($\bm{z}$) and alignment parameter ($\bm{R}$).
        The implicit neural representation (INR) predicts intensities and tissue probabilities for sampled coordinates, encoding subject-specific features in ($\bm{z}$) and general features in the shared weights $(\theta)$ of the INR. Optional conditioning ($\xi$), here LV volume, is concatenated to ($\bm{z}$). Simultaneously, rigid alignment of each subject to the common brain representation is learned through ($\bm{R}$). \textbf{(ii) Inference} generates a conditional spatio-temporal atlas. An atlas of age $t$ is represented by a weighted mean over the \emph{optimized} latent codes of training subjects around age $t$. Additional conditioning can modify the atlas by concatenating the desired LV volume to ($\bm{z}_t$). The final atlas in the desired modality and of time $t$ is generated through a forward pass of the INR modulated by ($\bm{z}_t$).
    }
    \label{fig:main_architecture}
\end{figure*}

\cinema~constitutes three phases: (Fig.~\ref{fig:main_architecture}-A) \textbf{Training} on a representative cohort to learn a general representation of the anatomy at hand. (Fig.~\ref{fig:main_architecture}-B) \textbf{Inference}, i.e., the generation, of an atlas with the desired attributes and characteristics. (C) \textbf{Test-time} adaptation to new subjects to perform downstream tasks like age prediction or tissue segmentation.

The following sections provide a theoretical perspective of the three phases.

\subsection{Training: Representation Learning}
\cinema, shown in Fig.~\ref{fig:main_architecture}, is designed as an auto-decoder approach \cite{Park2019, MetaSDF, Functa2022}, where the INR is modeled as multilayer perceptron (MLP) with sinusoidal activation functions as proposed in \cite{Siren}. The weights $\theta$ of the MLP are shared across all training data forcing the network to capture general, shared features only. Unlike prior approaches that use deformable registration to account for inter-subject variability, we encode subject variability in subject-specific parameters, termed \emph{latent codes}. 

\subsubsection{Latent Code}
Given $N$ training subjects, we condition the INR on latent codes \(\{\bm{z}_i\}_{i=1}^N\) via modulated layers, following the approach of \cite{Functa2022,Mehta, Dannecker2024CINA}. Specifically, to modulate layer $h$ of the INR, a latent code \(\bm{z} \in \mathbb{R}^D\) is transformed into scale (\(\bm{\alpha} \in \mathbb{R}^L\)) and shift (\(\bm{\beta} \in \mathbb{R}^L\)) parameters via a linear modulation layer :
\[
\begin{pmatrix} \bm{\alpha} \\ \bm{\beta} \end{pmatrix} = \bm{M}^h\bm{z} + \bm{\mu}^h,
\]
where \(\bm{M}^h \in \mathbb{R}^{2L \times D}\) and \(\bm{\mu}^h \in \mathbb{R}^{2L}\). The modulated output of INR layer $h$ is given by:
\[
\sin\left(\omega_0 \cdot \bm{\alpha} \cdot (\bm{W}^h\bm{x} + \bm{b}^h) + \bm{\beta}\right),
\]
where \(\bm{W}^h \in \mathbb{R}^{L \times L}\) and \(\bm{b}^h \in \mathbb{R}^L\) are the weights and biases of layer $h$ of the INR, \(\bm{x}\) is the 3D voxel coordinate, and \(\omega_0\) is a scaling factor introduced to better model high-frequency signals \cite{Siren}. Unlike \cite{Functa2022}, the modulation shift \(\bm{\beta}\) is not scaled by \(\omega_0\), as the latent code primarily captures temporal changes in brain dynamics, emphasizing low-frequency features to ensure a smooth latent representation. Note, while $\bm{z}$ is subject specific, $\bm{M}, \bm{\mu}, \bm{W}, \bm{b} \in \theta$ are shared across all training data. 

Following \cite{Park2019}, each subject in the training set is assigned a unique latent code \(\bm{z}\), initialized as \(\bm{z} \sim \mathcal{N}(0, 10^{-2})\).

\noindent \textbf{Spatial Latent Code.}
We have found that 1D latent codes, as used in CINA \cite{Dannecker2024CINA}, limit the spatial representation of complex anatomical features like the cortical gray matter of the brain. Here, we capitalize on the findings of Bauer et al. \cite{FunctaSpatial2023}, introducing spatial latent codes. We extend a subject's latent code to $\bm{z}\in\mathbb{R}^{\scriptscriptstyle D \times X_1 \times X_2 \times X_3}$, with $X_{1-3}$ defining the spatial dimensions of the latent code and $D$ defining the channels. To obtain the latent code at a specific coordinate \(\bm{x} \in \mathbb{R}^3\), we utilize trilinear interpolation over \(\bm{z}\). Whereas Bauer et al. \cite{FunctaSpatial2023} suggest employing convolutional neural networks (CNNs), we found trilinear interpolation, as in \cite{Mueller2022} to provide a more stable and computationally efficient alternative. Section~\ref{subsec:ablation_study} compares the two approaches more thoroughly.

\subsubsection{Subject Spatial Alignment}
\label{subsubsec:spatial_alignment}
\cinema~is designed to represent general, common features in the shared weights of the MLP. Unlike established methods, where inter-subject variability is pushed into the deformation fields during registration to achieve sharp representations, \cinema~encodes inter-subject variability into subject-specific latent codes. 
To avoid encoding irrelevant spatial information, we assign each subject $i$ a learnable rigid transformation composed of a rotation matrix $\bm{R}_i^{\text{rot}} \in \text{SO}(3)$ and a translation vector $\bm{t}_i \in \mathbb{R}^3$. This transformation is applied to each sampled coordinate $\bm{x} \in \bm{X}_i \subset \mathbb{R}^3$, where $\bm{X}_i$ denotes the set of voxel locations from subject $i$, as:

\[
\bm{R}_i(\bm{x}) = \bm{R}_i^{\text{rot}} \bm{x} + \bm{t}_i.
\]

The transformations $\{\bm{R}_i\}_{i=1}^N$ are optimized jointly with the latent codes and INR weights, ensuring consistent spatial alignment of all subjects to the shared anatomical representation.

\subsubsection{Training.}
Fig.~\ref{fig:main_architecture}-A outlines the training process. We maximize the joint log posterior over $N$ training subjects\textemdash{}each comprising (multimodal) brain MRI alongside tissue segmentation\textemdash{}optimizing the respective latent codes $\{\bm{z}_i\}_{i=1}^N$, the subject alignments $\{\bm{R}_i\}_{i=1}^N$, and the INR parameters $\theta$:

\begin{equation}
\resizebox{.9\hsize}{!}{$
\begin{aligned}
    \argmin_{\theta, \{\bm{z}_i\}_{i=1}^N, \{\bm{R}_i\}_{i=1}^N} 
    & \sum_{i=1}^N \Bigg[
    \sum_{\bm{x} \in \bm{X}_i} \mathcal{L}_{\text{MSE}}\left(
    f_{\theta}^{\text{img}}\left(\bm{R}_i(\bm{x}), \bm{z}_i(\bm{x})\right), \hat{I}_i(\bm{x})\right) \\
    & \quad + \mathcal{L}_{\text{CE}}\left(f_{\theta}^{\text{seg}}\left(\bm{R}_i(\bm{x}), \bm{z}_i(\bm{x})\right), \hat{C}_i(\bm{x})\right)
    \Bigg],
\end{aligned}
$}
\label{eq:log_posterior}
\end{equation}

with 

\begin{equation}
\resizebox{.9\hsize}{!}{$
\bm{z}_i(\bm{x}) = \begin{bmatrix} \text{Interp}(\bm{z}_i, \bm{x}) \\ \xi \end{bmatrix}, \hspace{0.5em} \bm{z}_i(\bm{x}) \in \mathbb{R}^{D+Q}, \hspace{0.5em} \bm{z}_i \in \mathbb{R}^{D \times X_1 \times X_2 \times X_3}.
$}
\label{eq:interp_latent}
\end{equation}

\(\text{Interp}(\bm{z}_i, \bm{x})\) performs trilinear interpolation on the latent code \(\bm{z}_i\) at voxel \(\bm{x}\), and \(\xi \in \mathbb{R}^Q\) represents explicit conditioning parameters (see Section~\ref{subsec:anat_conditioning}). \(\hat{I}_i(\bm{x})\) and \(\hat{C}_i(\bm{x})\) denote the ground-truth (multimodal) intensity and tissue label at voxel \(\bm{x}\) of subject i, while \(f_{\theta}^{\text{img}}\) and \(f_{\theta}^{\text{seg}}\) are the intensity and tissue probability predicted by the INR. The losses \(\mathcal{L}_{\text{MSE}}\) and \(\mathcal{L}_{\text{CE}}\) represent the mean squared error and cross-entropy terms, respectively. No explicit regularization was required for \(\bm{z}_i\), as the training process naturally enforces a compact representation.

\subsection{Inference: Atlas Generation}
\label{subsec:inf_atlas_gen}
After training, the INR has learned a general representation of the target domain, i.e., the brain, having pushed inter-subject variability into the latent codes $\{\bm{z}_i\}_{i=1}^N$. For perinatal brains, postmenstrual age (PMA) is expected to be the factor inducing the largest anatomical variability and is therefore also encoded in the latent codes. This is confirmed by plotting the first principal component of the latent codes against the subjects' PMA, revealing a high correlation as illustrated in Fig.~\ref{fig:main_architecture}-B. This age encoding allows us to generate a brain atlas for any target time point $t$ using a time regressed latent code $\bm{z}_t$ defined as:
\begin{equation}
    \bm{\bar{z}}_t = \sum_{i=1}^{N}w(t, t_i)\bm{z}_i,
\label{eq:mean_latent}
\end{equation}
Here, $\{t_i\}_{i=0}^N$ represent the GA of the $N$ training subjects normalized to $[-1, 1]$ and $w$ denotes a Gaussian kernel with mean $t$ and standard-deviation $\sigma$. A larger $\sigma$ yields more general but smoother atlases, whereas a smaller sigma introduces higher bias to individual subjects. We empirically set $\sigma \circumeq 0.5$ weeks, such that subjects within $\pm1$ week of $t$ contribute \textapprox 95\% of the total weight. Finally, by conditioning on the regressed latent code $\bm{\bar{z}}_t$, a forward pass defined as:

\begin{equation}
\label{eq:forward_pass}
\resizebox{.9\hsize}{!}{$
\textnormal{INR}_\theta(\bm{X}|\bm{\bar{z}}_t) = 
\sum_{\bm{x} \in \bm{X}}
\left(f_{\theta}^{img}\left(\bm{x}, \bm{\bar{z}}_t({\bm{x}})\right), f_{\theta}^{seg}\left(\bm{x}, \bm{\bar{z}}_t({\bm{x}})\right)\right),
$}
\end{equation}

yields the atlas with intensities and tissue probabilities for all voxels $\bm{x}\in$ $\bm{X}$. Here, $\bm{X}$ defines the 3D cartesian grid spanning our region of interest, i.e., the brain, in any desired resolution. \(\bm{\bar{z}}_t({\bm{x}})\) is defined in (\ref{eq:interp_latent}) and integrates any explicit conditions, e.g., the volume of the lateral ventricles (LV), into the atlas generation. Note, as we only perform a forward pass, all parameters and weights remain unchanged in the inference stage. 

\begin{figure*}
    \centerline{\includegraphics[width=\textwidth]{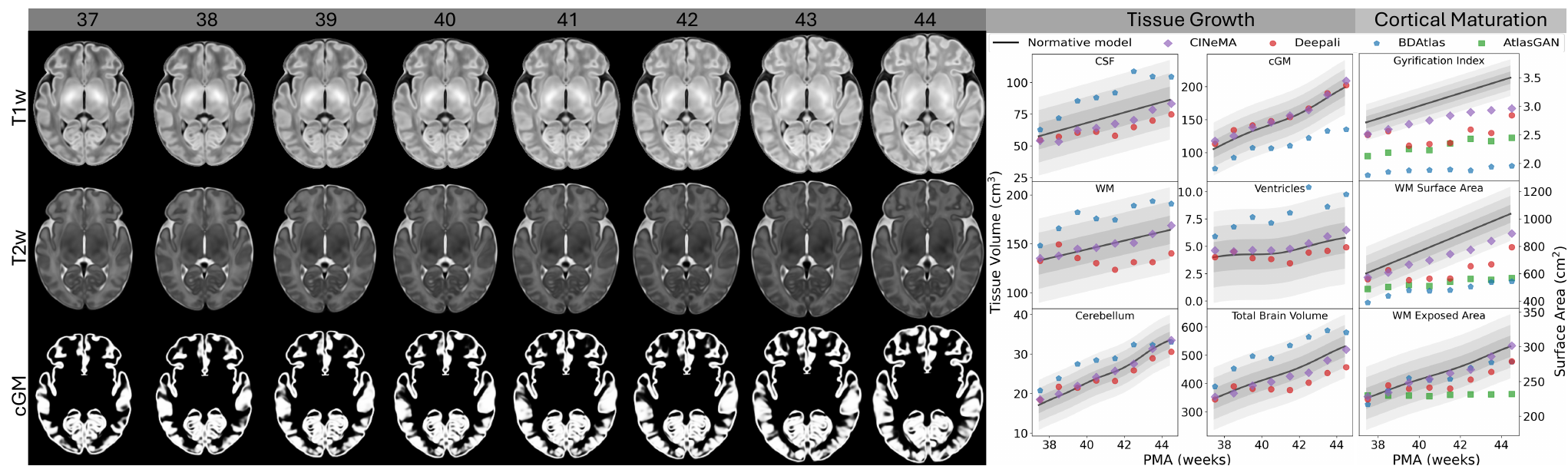}}
    \caption{(Left) Multimodal neonatal atlas for 37-44 weeks PMA generated with \cinema, depicting T1w, T2w, and cGM probability maps. (Right) Tissue growth and cortical maturation for atlases from \cinema~(purple) vs. baselines Deepali (red), BDAtlas (blue), and Atlas-GAN (green). Black: normative mean with $\pm$1--3 SDs computed with \cite{Dimitrova2021}. Gyrification index ($GI$) defined as $GI=\tfrac{WM_{surface}}{WM_{exposed}}$. Note: Atlas-GAN lacks volume analysis due to size normalization through affine pre-alignment, thus WM surface is not true to scale; $GI$ remains valid.}
    \label{fig:mm_neo_atlas}
\vspace{-0.4cm}
\end{figure*}

\begin{figure}[t]
    \centering
    \includegraphics[width=\columnwidth]{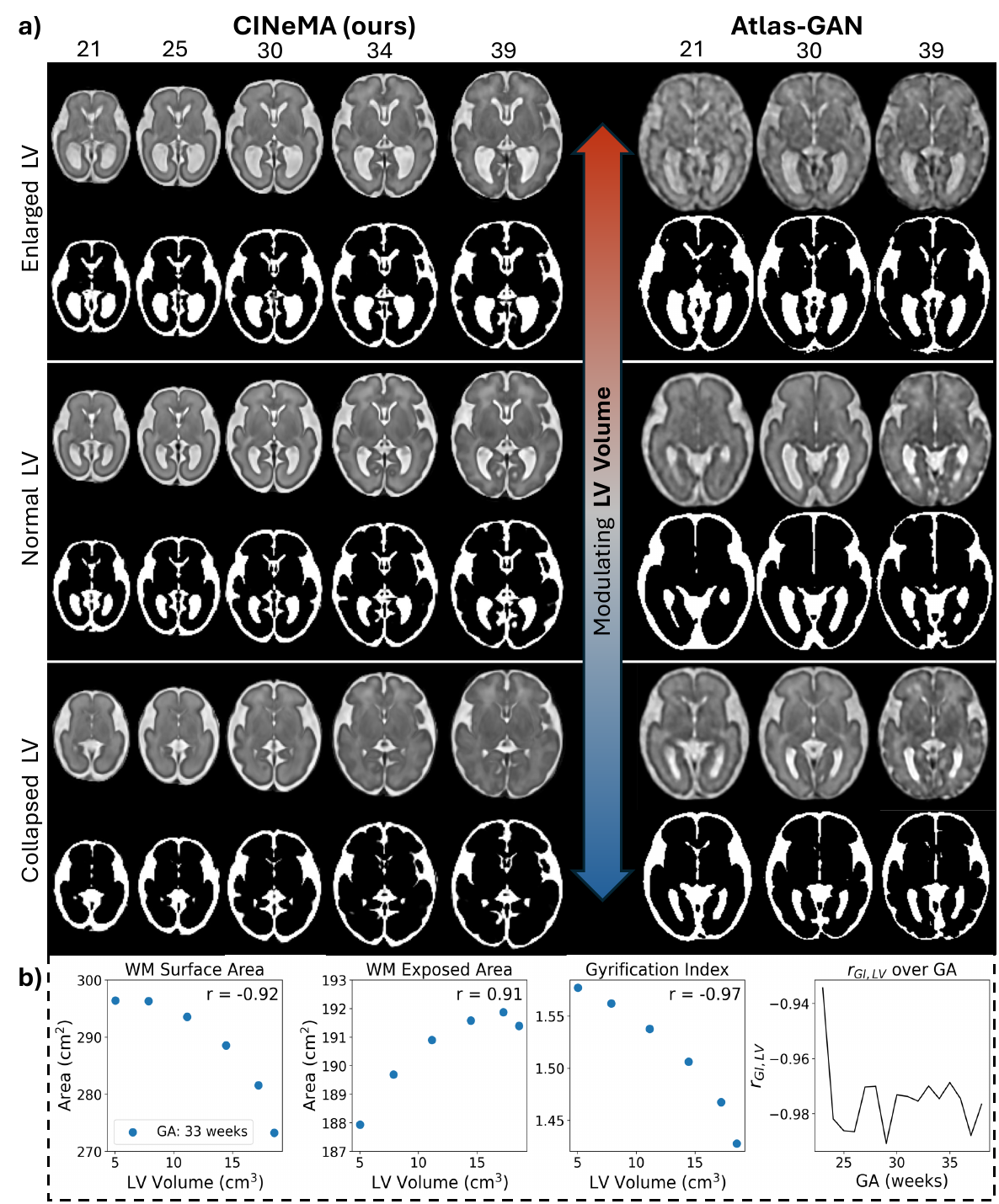}
    \vspace{-0.5cm}
    \caption{a) Conditional atlases for varying lateral ventricle (LV) volumes using \cinema~(left) and Atlas-GAN (right). Intensity and tissue maps for CSF and LV show enlarged (top), normal (middle), and collapsed (bottom) ventricles. Note: Atlas-GAN atlases are size-invariant due to affine pre-alignment. b) White matter (WM) surface area, WM exposed surface area, and gyrification index ($GI=\tfrac{WM_{surface}}{WM_{exposed}}$) of \cinema~atlas of week 33 conditioned on LV volume (x-axis). $GI$ decreases with increasing LV volume (Pearson Correlation of $r_{GI,LV}=-0.97$). Rightmost panel indicates a consistent trend across all ages.}
    \label{fig:LV_Atlas}
\end{figure}

Fig.~\ref{fig:mm_neo_atlas} and Fig.~\ref{fig:LV_Atlas} illustrate temporal atlases generated using this approach. The former shows an unconditioned neonatal atlas, while the latter demonstrates conditioning based on LV volume.

\subsection{Test-time: Adaptation to New Subjects}
\label{sec:ttime_adaptation}
To leverage an atlas for individualized analysis such as brain segmentation and subject-specific comparisons, it is necessary to adapt the general atlas to a new subject's anatomy. While traditional atlas-based approaches employ deformable registration techniques, 
we adapt \cinema~to a new subject $j$, by freezing the INR parameters \(\theta\), and optimizing a newly initialized latent code $\bm{z}_j \sim \mathcal{N}(0, 10^{-2})$ together with the zero-initialized rigid alignment parameter \(\bm{R}_j\), minimizing:

\begin{equation}
\label{eq:ttime_adaptation}
    \bm{z}_{j}, \bm{R}_j = \argmin_{\bm{z}, \bm{R}} \sum_{\bm{x}\in \bm{X}} \mathcal{L}\left(f_\theta^{img}\left(\bm{R}(\bm{x}), \bm{z}({\bm{x})}\right), \hat{I}_i\left(\bm{x}\right)\right),
\end{equation} 

where $\hat{I}_i(\bm{X})$ denote the intensities of subject $j$ and \(\bm{z}(\bm{x)}\) is defined by (\ref{eq:interp_latent}). The explicit conditions \(\xi_j\in\mathbb{R}^Q\) are randomly initialized as \(\xi \sim \mathcal{N}(0, 10^{-2})\) and also set as learnable. Note, we optimize on the image intensities only, thus do not require segmentation maps for test-time adaptation. To avoid overfitting, we use 10 \% of the subject intensities, randomly sampled within the brain mask, as holdout set to monitor the loss and apply early stopping. From the optimized latent code \(\bm{z}_{j}\), we can infer accurate tissue probability maps via a forward pass as defined by (\ref{eq:forward_pass}). Additionally, in a multimodal setup, we can use a single modality for latent code optimization, e.g., T2-weighted MRI (T2w), and, after optimization, infer the missing modality, e.g., T1w. We investigate this property more thoroughly in Section~\ref{sec:mm_atlas}. Finally, we can also utilize the explicit conditions \(\xi\), optimized during test-time adaptation, to gain additional insights to a subject. For example, Section~\ref{sec:conditioning_ba} demonstrates accurate birth age prediction of preterm neonates from the optimized explicit conditions.

\subsection{Conditioning on Anatomical Characteristics}
\label{subsec:anat_conditioning}
\cinema~encodes subject-specific features into individual latent codes $\{\bm{z}_i\}_{i=1}^N$, learning to separate them from shared features during training. Conditioning on specific anatomical characteristics during inference, however, poses the challenge of disentangling these characteristics from other encoded features. To address this, we introduce additional disjoint dimensions during training, $\xi \in \mathbb{R}^Q$, concatenated to the latent code $\bm{z}$. This approach enables disentangled conditioning on both discrete and continuous properties, such as birth age, or anatomical metrics (e.g., ACC or LV volume). After training, \cinema~can generate spatio-temporal atlases reflecting the desired condition, while supporting interpolation within and extrapolation beyond the training domain. During test-time adaptation, explicit conditions are jointly learned with the latent code (see Section~\ref{sec:ttime_adaptation}), enabling downstream tasks like birth age prediction (explored in Section~\ref{sec:conditioning_ba}).

\section{Experimental Setup}

\subsection{Data}
\label{sec:data}
\subsubsection{Neonatal Brain Data - dHCP}\hfill\\
\indent \textbf{Term Neonates.} We used 306 brain MRI scans from 304 term-born neonates (37–45 weeks PMA), 191 scans for training, 38 for validation, and 77 for testing. Imaging included T1- and T2-weighted MRI, following the protocol described in \cite{makropoulos2018developing}. All subjects had a radiology score of 1 indicating "normal appearance for age" \cite{Edwards2022}. Quality-controlled, segmented tissue maps were available for all scans \cite{makropoulos2018developing} and served as ground truth.

\textbf{Preterm Neonates.} For conditioning on birth age, we utilized T1- and T2-weighted MRI scans from 144 neonates with birth age ranging from 25 to 42 weeks PMA and scan age ranging from 37 to 45 weeks PMA. Of these, 52 scans were used for evaluation. Subjects had radiology scores of up to 3, indicating "incidental findings with unlikely clinical significance [...]" \cite{Edwards2022}. Quality checked tissue segmentation, obtained using dHCP-specific pipelines \cite{cordero2019dhcpRec, makropoulos2018developing}, were used as ground truth.

\subsubsection{Fetal Brain Data - dHCP}
We used 296 fetal brain MR images from 272 subjects, assigning 201 scans for training, 37 for validation, and 58 for testing. None of the test subjects was in the training set. GA ranged from 21 to 38 weeks. Detailed acquisition parameters are described in \cite{price2019dhcpACQ}. Super-resolution volumes with isotropic resolution of 0.5 mm were reconstructed using the SVR method of \cite{cordero2019dhcpRec}. Each scan included quality-controlled, automatically segmented tissue maps of brain regions \cite{makropoulos2018developing} serving as ground truth. Only T2-weighted scans were employed due to insufficient quality of T1-weighted scans. All scans underwent quality control checks.

\subsubsection{Fetal Brain Data - MarsFet}
The MarsFet dataset was constituted by retrospective access to MRI data acquired during routine clinical appointments at la Timone Hospital in Marseille between 2008 and 2021. 
This study was approved by the local ethical committee from Aix-Marseille University (N°2022-04-14-003). MRI were acquired using a T2-weighted half Fourier single shot turbo spin echo (HASTE) sequence on three MRI Siemens scanners (Skyra 3T, MagnetomVida 3T and SymphonyTim 1.5T). Super-resolution volumes with isotropic resolution of 0.5 mm were reconstructed using the NesVOR v0.2 method \cite{xu2023nesvor}. Close collaboration between medical doctors (N.G and M.M) enabled the identification of homogeneous subpopulations. To obtain segmentation ground truths, we employed a 2-step approach as detailed in \cite{Mihailov2024}: 1) We trained a 3D nnU-Net \cite{nnUNet} on the 50 youngest infants from the dHCP with available high quality segmentations; 2) We fine-tuned the nnU-Net on seven fetuses with manually labeled groundtruth. We applied nnU-Net to the MarsFet dataset, visually inspected all segmentations (M.D. and A.M.), and excluded low-quality data. Since the nnU-Net was trained on healthy data, segmentation failed for all subjects with \emph{severely} enlarged LV. Therefore, we performed manual corrections on the LV of five fetuses, serving us as groundtruth for the VM evaluation. The final cohort included 210 fetuses (22–38 weeks): 139 controls, 51 with isolated VM (46 non-severe, 5 severe), and 20 with isolated complete agenesis of the corpus callosum (ACC).




\subsection{Baselines}
\begin{enumerate}
    \item \textbf{BD-Atlas (fetal)\cite{Serag2012}:} 4D fetal brain atlas (23–37 weeks GA) built from T2w MRI of 80 fetuses. It employs free-form deformations \cite{rueckert1999nonrigid} and adaptive kernels for age-dependent atlas construction, with tissue probability maps for cerebrospinal fluid (CSF), cortical gray matter (cGM), lateral ventricles (LV), and brain hemispheres.
    \item \textbf{BD-Atlas (neonatal)\cite{Makropoulus2016}:} 4D neonatal brain atlas from 420 MR images, offering tissue probability maps for CSF, cGM, white matter (WM), LV, cerebellum (CBM), brain stem (BS), and deep gray matter (dGM). Both BD-Atlases are publicly available at \href{www.brain-development.org}{ brain-development.org}.
    \item \textbf{Deepali\cite{Schuh2024deepali}:} An open-source GPU-accelerated registration library. Following the approach of \cite{starck2024using}, using group-wise registration, we constructed temporal atlases from the same training data as \cinema, optimizing hyperparameters on validation sets. 
    \item \textbf{Atlas-GAN\cite{dey2021generative}:} A conditional generative adversarial atlas trained on the same data as \cinema. For pre-processing it requires affine alignment of the training subjects such as an affine atlas as reference. 
\end{enumerate}

\subsection{Pre-Processing}
Contrary to existing methods, \cinema~is designed to operate on raw data, requiring minimal data pre-processing and \emph{no} prior atlases for initial alignments. All subjects share the same orientation. Misalignments among subjects are left uncorrected, as \cinema~automatically learns rigid alignments during training (see Section~\ref{subsubsec:spatial_alignment}). We mask, skull-strip, and normalize brain intensities to $[0, 1]$.
The resolution-agnostic nature of INRs eliminates the need for resampling, preserving image details by avoiding interpolation.

\subsection{Training Parameters}
\label{subsec:training_params}
For all experiments and datasets, we utilized the same set of hyperparameters optimized on a validation set. The INR consisted of 5 hidden layers with 1024 units each, with modulation applied to hidden layers 1, 3, and 5. The latent code dimension was set to $256 \times 3 \times 3 \times 3$. The learning rate $l_r$ for the INR was set to $1e^{-4}$, the $l_r$ for the latent code was set to $5e^{-4}$, and the $l_r$ for the rigid subject alignment $\bm{R}$ was set to $7.5e^{-3}$. Batch size, defined as the number of coordinates sampled per iteration, was set to $25\,000$. Training was conducted for 1 epoch and completed in \textapprox 12 minutes. Subsequent atlas inference required \textapprox 3 seconds. Finally, test-time adaptation on new subjects was conducted by optimizing the latent code for $10$ epochs, where each epoch required \textapprox 5 seconds. Experiments were performed on an Nvidia A6000 GPU with a memory requirement of \textapprox 10 GB (a smaller batch size would further reduce memory requirements).

\section{Experiments and Results}
\subsection{Brain Segmentation}
\label{subsec:brain_segmentation}
Accurate brain segmentation is a critical step in neuroimaging analysis, enabling reliable tissue quantification and facilitating downstream tasks such as disease diagnosis and developmental analysis. This section evaluates the segmentation accuracy of \cinema~on both healthy and pathological subjects, and compares it against traditional and deep learning-based atlas methods. 

In atlas-based segmentation of perinatal brains, a suitable atlas (e.g., age-matched) is typically selected and deformably registered to the target subject using external tools like ANTs\cite{Avants2009}. In contrast, \cinema~generates subject-specific tissue probability maps by directly optimizing a latent code on the subject's image intensities (Eq.\ref{eq:ttime_adaptation}). After optimization, the adapted atlas with segmentation maps can be generated with a single forward pass (Eq.~\ref{eq:forward_pass}). This process eliminates the need for atlas selection and enables flexible adaptation to anatomical variability, which is particularly valuable in fetal and neonatal imaging where anatomy evolves rapidly and often deviates from normative templates.

For all methods, we evaluate image similarity between the subject and the registered or adapted atlas. Segmentation accuracy is assessed across multiple tissue classes: for neonates, CSF, cGM, WM, LV, CBM, BS, and dGM; for fetuses, CSF, cGM, LV, and brain hemispheres. As baseline methods do not support multimodal input, all experiments\textemdash{}including training and testing\textemdash{}were conducted using T2w MRI only. We evaluate performance across a diverse cohort, including healthy neonatal and fetal scans from the dHCP and both healthy and pathological fetal scans with severe ventriculomegaly from the MarsFet dataset.

\begin{figure}
\centerline{\includegraphics[width=\columnwidth]{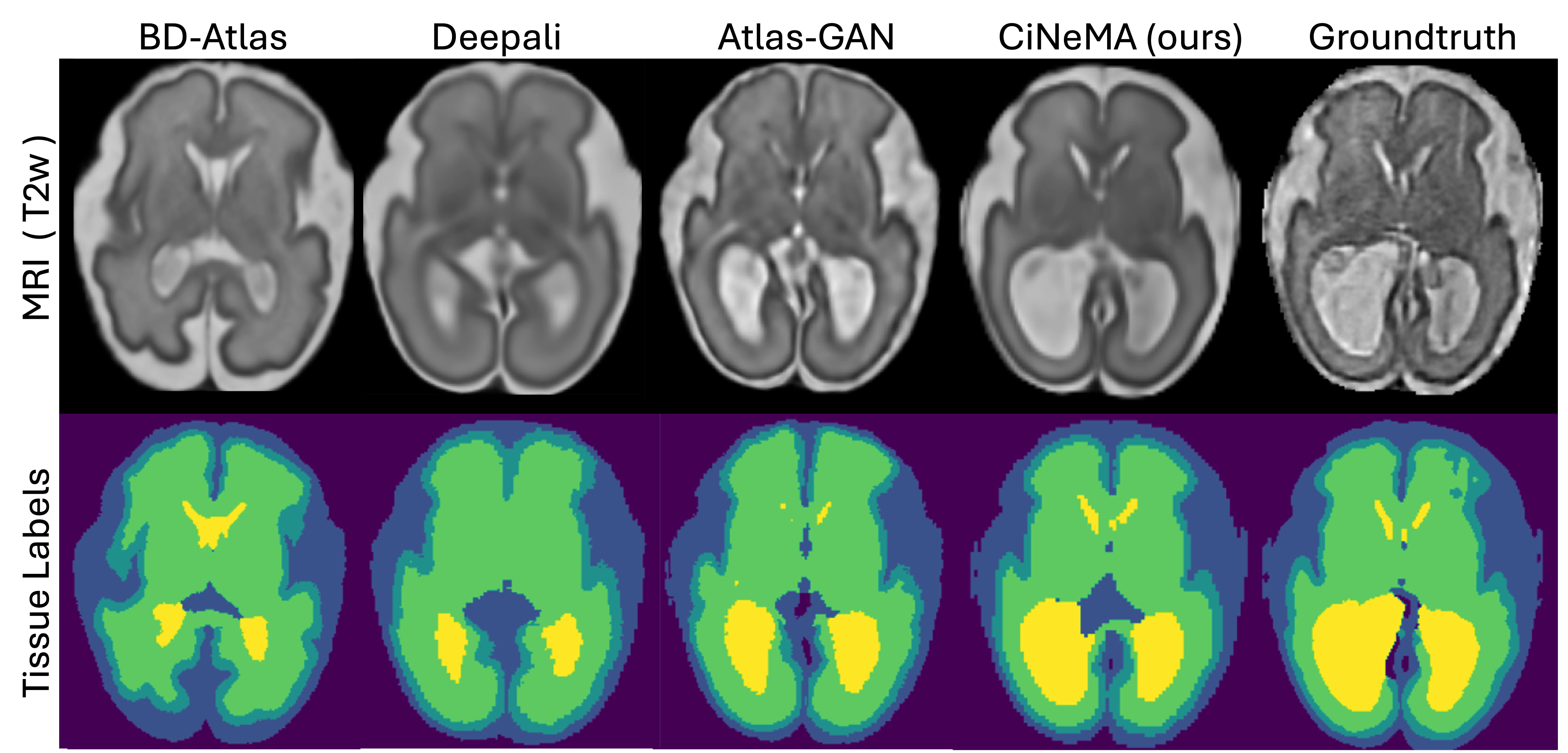}}
\caption{(Top) Atlases fitted to a new subject with severely enlarged LV (yellow label). All baselines were fitted using deformable registration \cite{Avants2009}. Atlas-GAN was conditioned on maximum possible LV size. \cinema~was fitted via latent code optimization (see Section~\ref{sec:ttime_adaptation}). (Bottom) Projected tissue labels. Right column shows the groundtruth.}
\label{fig:VM_qualitative}
\end{figure}

All experiment results are outlined in Table 1. \cinema~consistently demonstrates higher image similarity and more accurate tissue segmentation throughout all datasets including neonatal and fetal subjects. 
For fetal cases with severe ventriculomegaly, the differences are particularly striking. As expected, the traditional atlases, BDAtlas and Deepali, struggle to adapt the normative atlases to the abnormal anatomy (Fig.~\ref{fig:VM_qualitative}). The deep learning based Atlas-GAN, conditioned on LV volume, shows improved scores, but fails to fully capture the abnormal anatomy, lacking behind \cinema.

\begin{table}[ht]
\scriptsize
\caption{Brain segmentation and age prediction of individuals via atlas adaptation to T2w MRI of test subjects. We assess image similarity between subject and adapted atlas (PSNR, SSIM), segmentation accuracy of projected segmentation maps (DSC), and mean absolute error of predicted scan age (SA-MAE). MEAN $\pm$ STD over subjects; best in \textbf{bold}.}
\label{tab:results_segmentation}
\setlength{\tabcolsep}{2.75pt} 
\resizebox{\columnwidth}{!}{%
\begin{tabular}{l c c c c}
\toprule
Method & PSNR $\uparrow$ & SSIM $\uparrow$ & DSC $\uparrow$ & SA-MAE $\downarrow$\\
\midrule
\multicolumn{5}{c}{dHCP neonates (term) - 77 subjects } \\
\midrule
Atlas-GAN\cite{dey2021generative} & $\bm{23.75^* \pm 0.88}$ & $0.75 \pm 0.02$ & $0.78 \pm 0.01$ & N/A\\
BDAtlas\cite{Makropoulus2016} & $19.55 \pm 1.05$ & $0.54 \pm 0.12$ & $0.71 \pm 0.03$ & $4.48 \pm 1.40$\\
Deepali\cite{Schuh2024deepali} & $17.88 \pm 0.80$ & $0.71 \pm 0.01$ & $0.79 \pm 0.01$ & $1.81 \pm 1.49$ \\
\cinema~(ours) & $23.43 \pm 0.55$ & $\bm{0.82^* \pm 0.01}$ & $\bm{0.83^* \pm 0.03}$ & $\bm{0.96^* \pm 0.76}$ \\ 
\midrule
\multicolumn{5}{c}{dHCP fetal (neurotypical) - 58 subjects} \\
\midrule
Atlas-GAN\cite{dey2021generative} & $23.40 \pm 2.15$ & $0.84 \pm 0.05$ & $0.80 \pm 0.06$ & N/A\\
BDAtlas\cite{Serag2012} & $16.83 \pm 1.16$ & $0.46 \pm 0.11$ & $0.71 \pm 0.05$ & $1.50 \pm 1.22$ \\
Deepali\cite{Schuh2024deepali} & $15.69 \pm 1.23$ & $0.77 \pm 0.05$ & $0.83 \pm 0.07$ & $1.14 \pm 0.73$ \\
\cinema~(ours) & $\bm{24.67^* \pm 2.25}$ & $\bm{0.88^* \pm 0.07}$ & $\bm{0.85 \pm 0.08}$ & $\bm{0.88 \pm 0.95}$ \\
\midrule
\multicolumn{5}{c}{MarsFet (neurotypical) -  22 subjects} \\
\midrule
Atlas-GAN\cite{dey2021generative} & $20.40 \pm 2.82$ & $0.76 \pm 0.06$ & $0.78 \pm 0.04$ & N/A\\
BDAtlas\cite{Serag2012} & $15.65 \pm 1.97$ & $0.45 \pm 0.08$ & $0.68 \pm 0.04$ & $1.55 \pm 1.23$\\
Deepali\cite{Schuh2024deepali} & $14.32 \pm 1.58$ & $0.57 \pm 0.03$ & $0.40 \pm 0.02$ & $0.95 \pm 1.15$ \\
\cinema~(ours) & $\bm{24.38^* \pm 0.50}$ & $\bm{0.85^* \pm 0.02}$ & $\bm{0.83^* \pm 0.04}$ & $\bm{0.64^* \pm 0.71}$\\ 
\midrule
\multicolumn{5}{c}{MarsFet (severe ventriculomegaly) - 5 subjects} \\
\midrule
Atlas-GAN\cite{dey2021generative} & $15.65 \pm 3.15$ & $0.64 \pm 0.09$ & $0.83^\dagger \pm 0.05$ & N/A\\
BDAtlas\cite{Serag2012} & $11.69 \pm 1.77$ & $0.27 \pm 0.07$ & $0.33^\dagger\pm 0.03$ & $2.80 \pm 2.14$\\
Deepali\cite{Schuh2024deepali} & $11.23 \pm 1.29$ & $0.49 \pm 0.04$ & $0.20^\dagger \pm 0.02$ & $1.80 \pm 1.33$ \\
\cinema~(ours) & $\bm{23.44^* \pm 0.70}$ & $\bm{0.83^* \pm 0.03}$ & $\bm{0.89^{\dagger*} \pm 0.02}$ & $\bm{1.40 \pm 1.20}$\\ 
\bottomrule
\bottomrule 
  \multicolumn{5}{l}{%
    \scriptsize
    \shortstack[l]{%
      \\[0.05ex] 
      *Value significantly better than the second best ($p<0.05$, paired t‐test).\\
      $\dagger$ Lateral ventricles only.
    }
  } \\ 
\end{tabular}
}
\end{table}

Note in the above experiments, \cinema~was \emph{not} explicitly conditioned on lateral ventricle (LV) volume. Anatomical variability, including ventricular enlargement, was captured implicitly through the subject-specific latent codes $\bm{z}$. Through explicit conditioning on LV volume, we have not observed any benefits in the above experiments.  
The primary objective of explicit conditioning is to effectively disentangle specific anatomical properties from latent space. This disentanglement is what enables us to generate atlases that specifically model a certain anatomy, abnormality, or pathology as demonstrated in the following section.

\subsection{Modeling of Abnormal Populations}
\label{sec:conditioning}
In clinical and developmental neuroimaging, certain populations—such as preterm neonates or fetuses with structural brain abnormalities—exhibit anatomy that differs substantially from normative patterns. Modeling such deviations requires flexible and interpretable frameworks that can explicitly account for known clinical or biological variables. 

This section evaluates \cinema's ability to explicitly model abnormal brain populations by conditioning on relevant variables. We demonstrate this across three distinct scenarios where we condition on: (1) postmenstrual age (PMA) at birth for neonates, which is known to influence early brain development; (2) lateral ventricle (LV) volume of fetal brains, a key biomarker in ventriculomegaly (VM); and (3) agenesis of the corpus callosum (ACC) in fetal brains, a congenital condition captured via binary labels, i.e., the presence or complete absence of the corpus callosum. We trained \cinema~for each of these scenarios following the procedure described in Section~\ref{subsec:anat_conditioning}. Together, the experiments highlight the model’s capacity to generate anatomically coherent and clinically meaningful atlases along continuous and discrete axes of variation.

\subsubsection{Modeling Birth Age}
\label{sec:conditioning_ba}
\begin{figure}[ht]
\centerline{\includegraphics[width=\columnwidth]{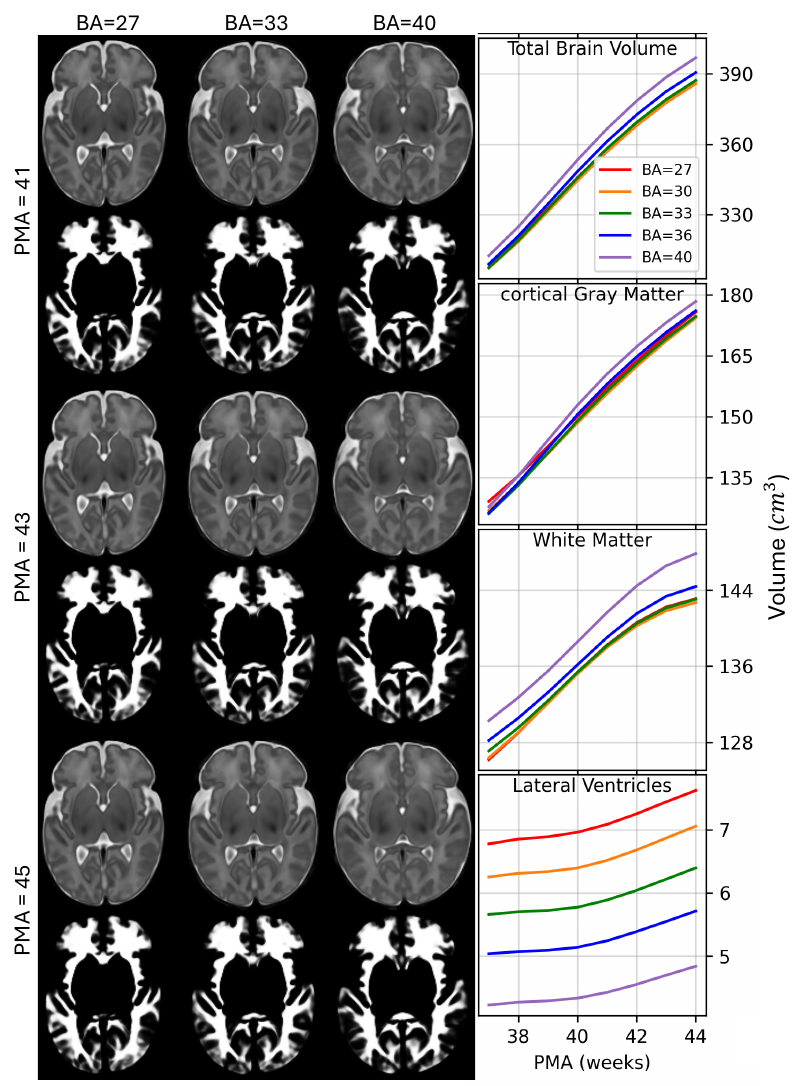}}
\caption{(Left) Spatio-temporal atlas with WM probability maps explicitly conditioned on birth age (BA) in weeks. Columns show different BA. Rows show scan age in weeks (PMA). (Right) Brain tissue volumes over scan age for different BA.} 
\label{fig:BA_conditioning}
\end{figure}

PMA at birth strongly impacts brain development and has differing effects on the volume of various brain regions. Previous studies such as \cite{PretermBrainDevel} reported impaired brain development in preterm neonates, defined as birth before 37 weeks GA, compared to term born infants. We trained \cinema~on a cohort of 92 term and preterm neonates from the dHCP dataset (25 to 42 weeks PMA at birth) while explicitly conditioning on birth age. After training, we generated spatio-temporal atlases conditioned on birth age, as illustrated in Fig.~\ref{fig:BA_conditioning}. For earlier birth age, we observe a decrease in total brain volume, cortical gray matter (cGM) and white matter (WM) tissue volumes. The volume of the LV, on the other hand, increases with earlier birth age. These tissue growth trajectories are in line with findings reported in literature \cite{Makropoulus2016, dHCPNeoPreterm2022}. Moreover, test-time adaptation on 52 term and preterm neonates yields an accurate birth age prediction of $1.51 \pm 0.16$ weeks mean absolute error (MAE) which compares well to birth age predictions reported in literature. For example, the authors of \cite{dHCPNeoPreterm2022} reported an accuracy of $2.21$ weeks MAE on term and preterm neonates of the same dataset. 

\subsubsection{Modeling Ventriculomegaly}
To adequately model ventriculomegaly (VM) in fetal brains, we found it particularly effective to condition on the volume of the lateral ventricles (LV). Whereas diagnosis of VM usually involves the width of the atrial diameter, there also exists a strong linear relationship with the LV volume \cite{Ma_VM_2019}. We trained \cinema~on 46 fetuses with varying degree of VM and 139 controls from the MarsFet dataset. During training, we extract the LV volume from the corresponding label maps, normalize it by the total brain volume and rescale it to $[-1 , 1]$. Next, we add the LV volume, denoted by \(\xi\), as disjoint, i.e., \emph{static}, dimension to a subject's latent code \(\bm{z}\) as defined by (\ref{eq:interp_latent}). After training, we can generate a spatio-temporal atlas with any desired LV volume by modulating the degree along the explicit dimension 
\(\xi\) as shown in Fig.~\ref{fig:main_architecture}-B and Fig.~\ref{fig:LV_Atlas}. Unlike Atlas-GAN, which struggles with combinations of GA and LV volume rarely or never seen during training, \cinema~generates high-definition brain atlases for these cases. Disjointly learned representations of LV and other brain anatomy enables smooth transitions along temporal and LV volume dimensions, letting \cinema~faithfully model age and LV without direct training examples. Sections \ref{subsubsec:modulation_anat} and \ref{subsubsec:ip_of_subjects} further validate this, demonstrating smooth interpolations between explicit conditions and latent codes.
Notably \cinema also captures intricate relationships between enlarged ventricles and cortical folding\cite{Benkarim2018}, showing a decrease in folding with an increase of LV volume for a fixed age (Fig.~\ref{fig:LV_Atlas}-b).
A comprehensive view of the atlas is provided as video in the supplementary material.

\subsubsection{Modeling Agenesis of the Corpus Callosum}
While LV volume represents a continuous condition, we can also use discrete labels for anatomical conditioning. Here we evaluate explicit conditioning on the presence or absence of the corpus callosum (CC). The CC is the main link between the two brain hemispheres and is made up of nerve fibers, i.e., white matter (WM). The absence of the CC represents a congenital disorder referred to as agenesis of the corpus callosum (ACC). For CC conditioning, we model the explicitly conditioned domain as $-1$ for neurotypical brains, i.e., presence of CC, and $1$ for brains with complete ACC. The training was conducted on 20 cases with complete ACC and 65 neurotypical subjects across the age of 22 to 32 weeks GA from the MarsFet dataset. After training, \cinema~faithfully generates generic brains with complete ACC, neurotypical brains, and also brains in-between, i.e., between $-1$ and $1$, representing CC conditions qualitatively similar to hypoplasia (see Fig.~\ref{fig:ACC_Atlas}). 
A detailed view of the atlas is provided as video in
the supplementary material.
\begin{figure}
\centerline{\includegraphics[width=\columnwidth]{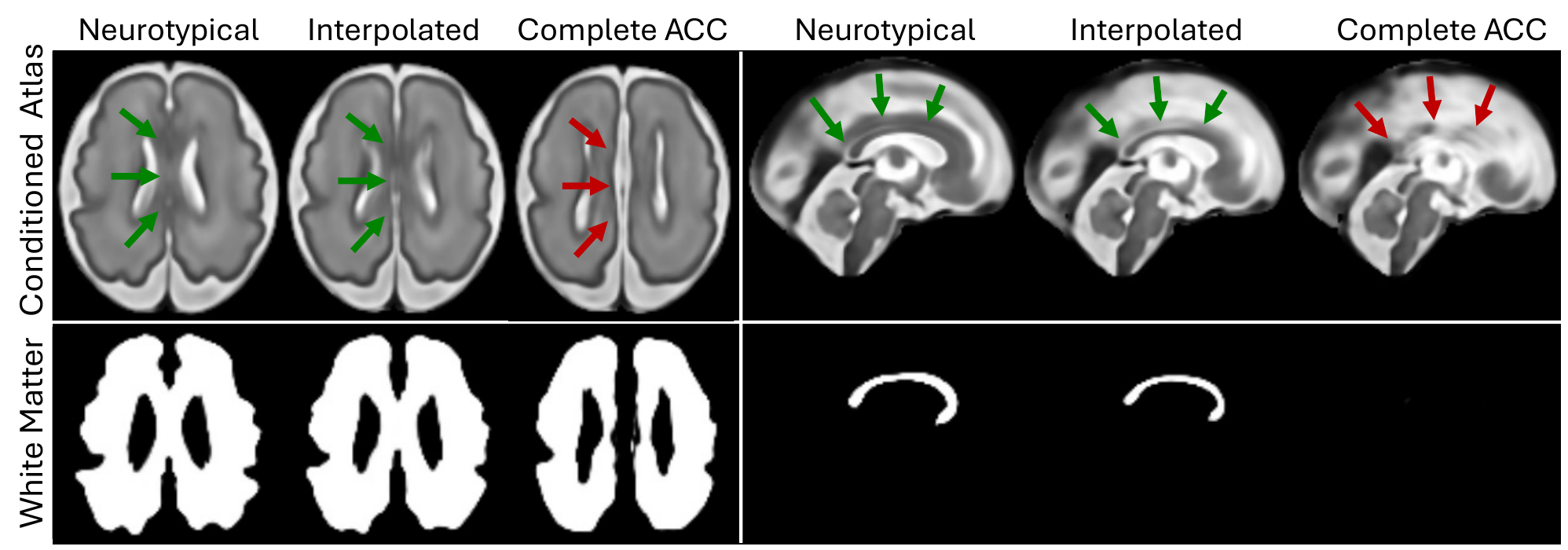}}
\caption{Fetal brain atlas at 27 weeks GA, conditioned on ACC with corresponding white matter (WM) probability maps, generated by \cinema. Axial (left) and sagittal (right) views show a normal CC (left), complete ACC (right), and an interpolated state resembling CC hypoplasia (middle).} 
\label{fig:ACC_Atlas}
\end{figure} 

\subsection{Multi-Modal Atlas}
\label{sec:mm_atlas}

\cinema~supports both mono and multimodal input. This study evaluates T1w and T2w MRI scans, however, there is no restriction on the number or types of modalities used. During training, all modalities are concatenated to a multi-channel image. After training, the spatio-temporal atlas can be queried in any of the trained modalities as illustrated in Fig.~\ref{fig:mm_neo_atlas}. Representation of multiple modalities increases the versatility of the atlas, but also improves the learned representation. Indeed, the ablation study (see Table~\ref{tab:ablation_study}) shows that multimodality induces small improvements in scan age prediction and significant improvement in birth age prediction of term and preterm neonates. In addition, the  learned multimodal representation opens up interesting generative potential of translating data from one modality to another.

\subsection{Additional Downstream Tasks}
We evaluate two additional downstream tasks: (1) modality translation, and (2) scan age prediction. 
\label{sec:downstream_tasks}

\subsubsection{Modality Translation}
\begin{figure}
\centerline{\includegraphics[width=\columnwidth]{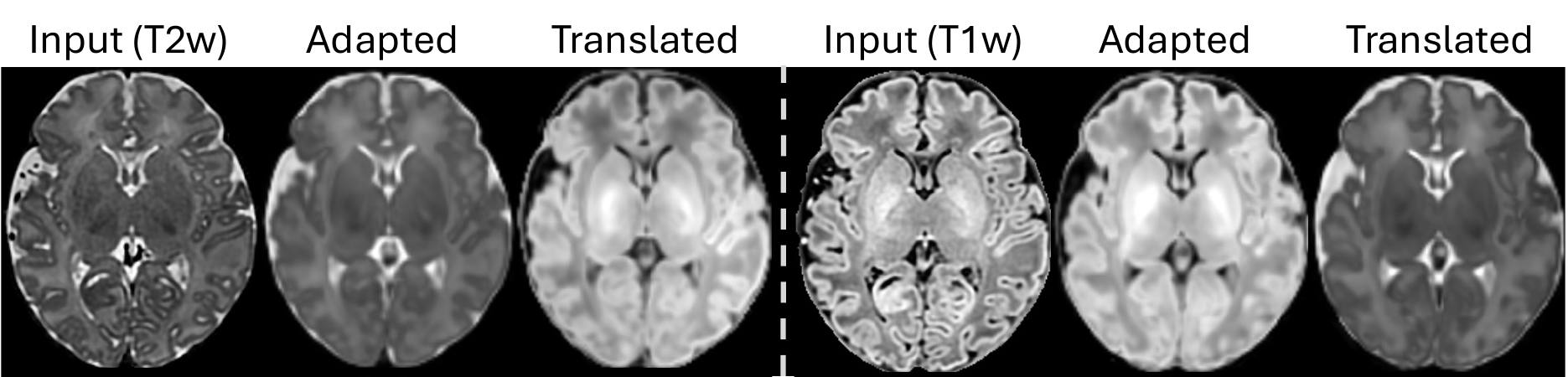}}
\caption{Translation/prediction to/of unseen modalities during test-time adaption on new subjects. (left) Atlas is adapted to a new T2w scan, with translation to T1w. (right) Adapted to T1w scan with translation to T2w.}
\label{fig:mod_translation}
\vspace{-0.4cm}
\end{figure}

During multimodal training, \cinema~automatically learns to translate between image modalities. Thus, at test-time, we can optimize a new subject's latent code $z$ on one modality to infer the corresponding representations in unseen modalities. As shown in Fig.~\ref{fig:mod_translation}, this translation achieves high fidelity, with mean similarity metrics of $25.31\pm 1.3$ PSNR and $0.90\pm0.02$ SSIM between adapted and translated T1w images, and $28.36\pm 2.48$ PSNR and $0.92\pm0.02$ SSIM for T2w images for 20 term neonates from the dHCP dataset (38 - 45 weeks PMA).

\subsubsection{Scan Age Prediction}
\label{subsec:scan_age_pred}
In addition to segmentation, \cinema~provides precise age estimates for test subjects by analyzing the optimized latent codes via Neighborhood Component Analysis, using the latent codes of the training subjects as reference. Since all baselines lack direct age estimation, we employed linear regression models for each baseline, utilizing their atlas brain volumes as reference points to fit the model. Table~\ref{tab:results_segmentation} highlights consistently higher accuracy of \cinema~in age prediction compared to the baselines across all experiments. The improvement is particularly pronounced in neonatal subjects, where brain volume alone proves to be a less reliable age predictor due to the slowed rate of brain growth during this stage of development. Note, as Atlas-GAN required affine pre-alignment of training subjects, age regression by brain volume was not possible.

\subsection{Generative Properties}
\label{subsec:gen_properties}
\cinema~encodes subject properties into a compact latent code $\bm{z}$ enabling interpolation between subjects and the generation of new data. The same applies to explicitly conditioned attributes, further expanding its generative potential. The following sections explore this in detail.
\subsubsection{Modulation of Specific Anatomical Features}
\label{subsubsec:modulation_anat}
\begin{figure}
\centerline{\includegraphics[width=\columnwidth]{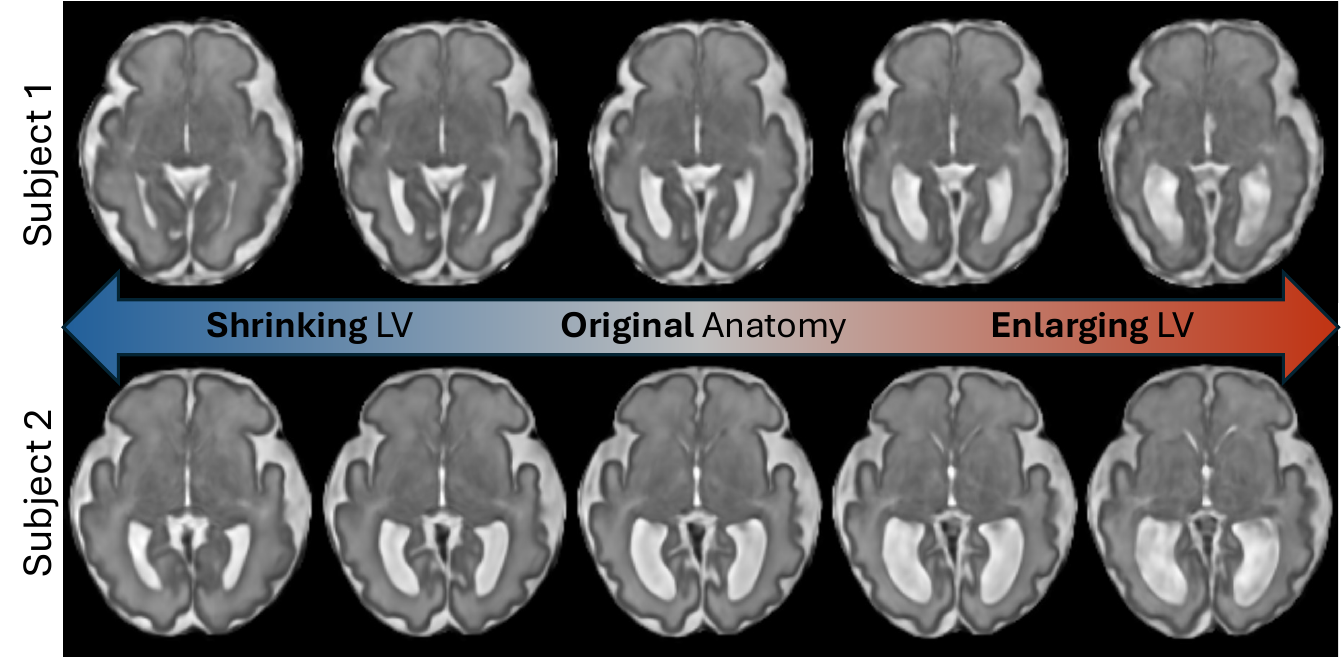}}
\caption{Modulation of explicitly conditioned LV volume, of \emph{individual} subjects. Original subject in the middle. \cinema~can shrink or enlarge the LV of subjects, preserving the individual anatomy.}
\label{fig:LVIV_modulation}
\end{figure} 

Explicit conditioning during training ensures a disentangled encoding between conditioned anatomy and other brain structures. This allows modulation of anatomical features for atlases (Fig.~\ref{fig:LV_Atlas}) but also for individual subjects (Fig.~\ref{fig:LVIV_modulation}). For individual subjects, conditioning on LV volume enables controlled enlargement or shrinkage without losing subject-specific anatomical details. This demonstrates \cinema's ability to extrapolate conditioned anatomy leveraging information learned from other training subjects, potentially of different age, without collapsing to a generic brain representation. This capability also serves as \emph{anatomically informed} data augmentation.

\subsubsection{Interpolation Between Latent Codes}
\label{subsubsec:ip_of_subjects}
\begin{figure}
\centerline{\includegraphics[width=\columnwidth]{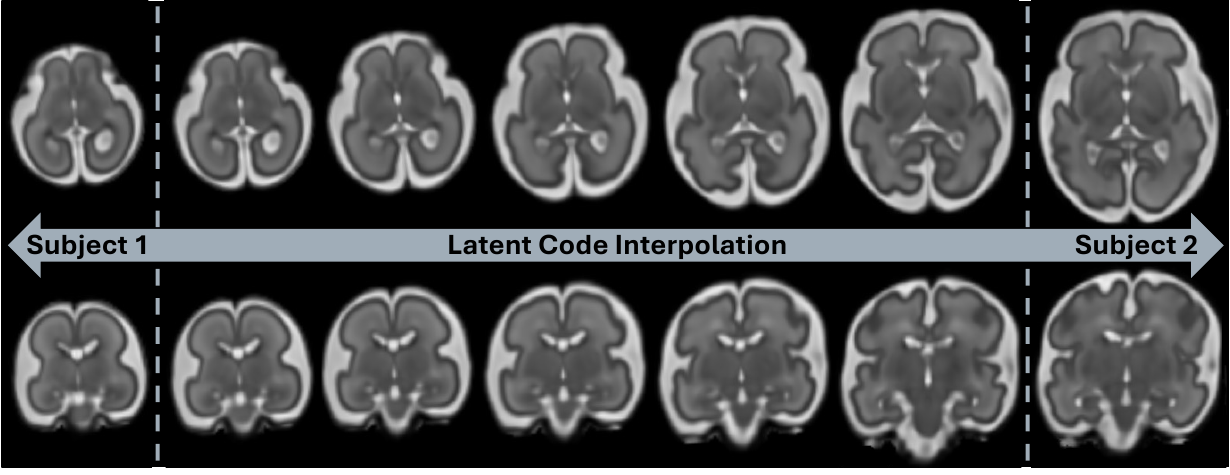}}
\caption{Image reconstruction of latent codes interpolated between two distant subjects (22 weeks and 31 weeks GA).}
\label{fig:lat_code_ip}
\end{figure}

\cinema~encodes subject-specific features into latent codes \(\{\bm{z}_i\}_{i=1}^N\), which form a compact latent space with smooth and continuous transitions. Notably, the extension from $1D$ to $4D$ spatial latent codes, considerably increasing subject-specific parameters, maintains this continuity without any indication of overfitting. This is evidenced by the smooth interpolations required for spatio-temporal atlas generation (Fig.~\ref{fig:mm_neo_atlas}) and interpolations between latent codes of age-distant subjects (Fig.~\ref{fig:lat_code_ip}), producing realistic anatomical representations with artifact-free transitions.

\subsection{Ablation Study}
\label{subsec:ablation_study}
In this section, we evaluate the architectural modifications of \cinema~compared to its predecessor, CINA \cite{Dannecker2024CINA}.  
These include multimodal input integration, the extension from 1D latent codes $\bm{z}^{256}$ to spatial latent codes $\bm{z}^{256 \times 3^3}$, 
such as the introduction of learnable rigid subject alignment $\bm{R}$. Below and in Table~\ref{tab:ablation_study}, we detail the impact of these changes.

\subsubsection{Multimodal Input}
The inclusion of multimodal input (T1w and T2w MRI) shows significant improvements over monomodal T1w in dice score and, over monomodal T2w, for birth age prediction (Table~\ref{tab:ablation_study}). T2w provides superior contrast and quality for neonate imaging explaining the observed benefit of adding T2w alongside T1w.

\subsubsection{Rigid Subject Alignment}
Integrating learnable rigid subject alignment $\bm{R}$ significantly improves performance across all metrics (Table~\ref{tab:ablation_study}). By encoding subjects' spatial positions in $\bm{R}$ instead of $\bm{z}$, \cinema~maintains spatial consistency across latent codes during training. This ensures better segmentation accuracy and more informative latent representations.

\subsubsection{Spatial Latent Codes}
As shown in Table~\ref{tab:ablation_study}, the extension to spatial latent codes $\bm{z} \in \mathbb{R}^{256 \times 3^3}$ yields significant improvements in dice score compared to 1D latent codes $\bm{z} \in \mathbb{R}^{256}$.
Even with matching parameter count of latent codes $\bm{z} \in \mathbb{R}^{6912}$, spatial codes outperformed, highlighting their capacity to represent complex anatomical structures.

\subsubsection{Latent Code Spatial Interpolation}
For our experiments we employ spatial latent codes with trilinear interpolation. The extension to spatial interpolation with a convolutional layer, as proposed in \cite{FunctaSpatial2023}, did not yield significant benefits even when expanding the code size from $3^3$ to $5^3$. On the contrary, the model performed significantly worse for birth age prediction. 

\subsubsection{Explicit vs Implicit Conditioning}
As discussed in Section~\ref{subsec:brain_segmentation}, the primary objective of explicit conditioning is to disentangle specific anatomical features from the latent space. While Section~\ref{sec:conditioning} demonstrates the effective disentanglement qualitatively, it is also reflected by improved predictive power of the conditioned variable, e.g., birth age, on new subjects.
When adapting to a new subject, both the latent code and the condition variable are randomly initialized and jointly optimized. After convergence, the optimized condition reflects \cinema's estimate of that variable. Table~\ref{tab:ablation_study} compares birth age (BA) prediction accuracy using explicit vs. implicit conditioning (last row). In the explicit case, BA is inferred directly from the optimized condition. In the implicit case, BA is estimated using neighborhood component analysis, similar to scan age, as described in Section~\ref{subsec:scan_age_pred}. Explicit conditioning achieves significantly better accuracy, indicating more effective disentanglement of the  anatomical feature.  

To summarize, this ablation study highlights the effectiveness of \cinema's architectural enhancements, supporting their integration into the framework for improved performance across multiple tasks.

\begin{table}[ht]
\scriptsize
\caption{Ablation study on multimodal input, rigid alignment ($\bm{R}$), explicit conditioning (E/C), spatial versus $1D$ latent codes ($\bm{z}$), and latent code spatial interpolation via a convolutional network (CNN) with kernel size $k$ or trilinear (\xmark). Metrics include mean dice scores (DSC), and mean absolute error of scan age (SA-MAE) and birth age (BA-MAE) in weeks. Mean metrics over five random seeds. Best in \textbf{bold}.}
\label{tab:ablation_study}
\setlength{\tabcolsep}{2.75pt} 
\resizebox{\columnwidth}{!}{%
\begin{tabular}{c c c c c c | c c c}
\toprule
\multicolumn{6}{c|}{\textbf{Configuration}} & \multicolumn{3}{c}{\textbf{Metrics}} \\
\midrule
E/C & T1w & T2w & $\bm{R}$ & $\bm{z}$ & CNN & DSC $\uparrow$ & SA-MAE $\downarrow$ & BA-MAE $\downarrow$\\
\midrule
\cmark &\cmark & \cmark & \cmark & $256$ & \xmark & $0.77^* \pm 0.01$ & \mbox{\bm{$1.02\; \pm 0.16$}} & $2.06^* \pm 0.25$ \\
\cmark &\cmark & \cmark & \cmark & $6912$ & \xmark & $0.78^* \pm 0.02$ & $2.34^* \pm 0.19$ & $3.73^* \pm 0.31$ \\
\cmark &\cmark & \cmark & \xmark & $256 \times 3^3$ & \xmark & $0.82^* \pm 0.01$ & $1.76^* \pm 0.12$ & $1.74^* \pm 0.09$\\
\cmark &\cmark & \xmark & \cmark & $256 \times 3^3$ & \xmark & $0.80^* \pm 0.00$ & $1.33^* \pm 0.07$ & $1.56^* \pm 0.16$  \\
\cmark &\xmark & \cmark & \cmark & $256 \times 3^3$ & \xmark & $0.84\;\; \pm 0.01$ & $1.41^* \pm 0.17$ & $1.71^* \pm 0.08$ \\
\rowcolor{gray!20} 
\cmark &\cmark & \cmark & \cmark & $256 \times 3^3$ & \xmark & $0.84\;\; \pm 0.01$ & $1.20\;\; \pm 0.20$ & \mbox{\bm{$1.50\; \pm 0.14$}} \\
\cmark &\cmark & \cmark & \cmark & $256 \times 3^3$ & $k=3$ & $0.83^* \pm 0.00$ & $1.63^* \pm 0.28$ & $1.58^* \pm 0.13$ \\
\cmark &\cmark & \cmark & \cmark & $256 \times 5^3$ & $k=3$ & \mbox{\bm{$0.85\; \pm 0.01$}} & $1.23\;\; \pm 0.07$ & $1.89^* \pm 0.29$ \\
\cmark &\cmark & \cmark & \cmark & $256 \times 5^3$ & $k=5$ & $0.84\;\; \pm 0.01$ & $1.92\;\; \pm 0.83$ & $1.86^* \pm 0.15$ \\
\xmark &\cmark & \cmark & \cmark & $256 \times 3^3$ & \xmark & $0.84\;\; \pm 0.01$ & $1.29\;\; \pm 0.31$ & $3.29^* \pm 1.01$ \\
\bottomrule
\bottomrule 
  \multicolumn{9}{l}{%
    \scriptsize
    \shortstack[l]{%
      \\[0.05ex] 
      Shaded row indicates model configuration used in this study.\\
      *Value significantly worse than the best ($p<0.05$, paired t-test).
    }
  }
\end{tabular}
}
\end{table}

\begin{figure}
\centerline{\includegraphics[width=\columnwidth]{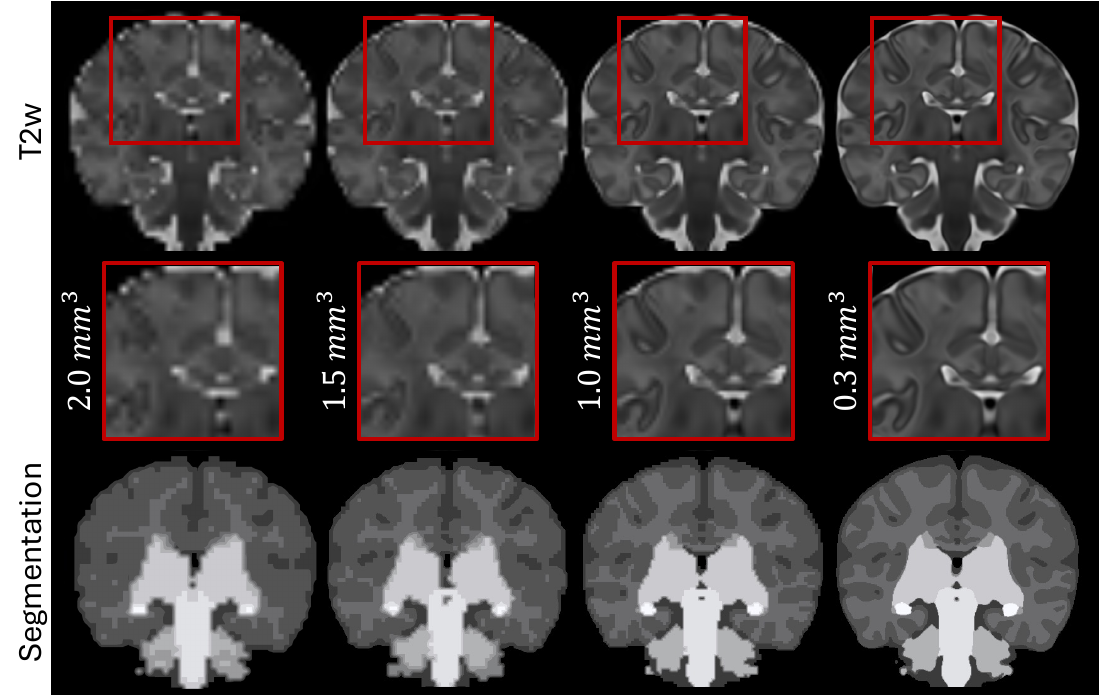}}
\caption{Neonatal atlas at 42 weeks PMA generated with \cinema~at four resolutions (2.0--0.3mm$^3$) using the same model. Anatomical consistency is preserved up to partial volume effects. Memory footprints of full temporal atlas (37-45 weeks): 1.0mm$^3$ \textapprox 40MB, 0.3mm$^3$ \textapprox 1.2GB, \emph{all} model parameters just 30MB.}
\label{fig:multi_res_atlas}
\end{figure} 

\section{Discussion}

\subsection{Latent Representation and Conditioning}
\subsubsection*{Disentanglement of Shared and Individual Features}
\cinema~utilizes an auto-decoder framework \cite{Functa2022, Mehta, Park2019} that inherently promotes disentanglement between shared anatomical features, encoded within the shared INR parameters $\theta$, and individual anatomical variations captured by the latent codes $\bm{z}$. Through this setup, the INR cannot effectively encode individual variations without compromising reconstruction performance, i.e., the optimization target defined in (\ref{eq:log_posterior}). In an extreme scenario without latent codes, the INR would learn to reconstruct the population mean for all subjects, failing to reconstruct any subject-specific details.

Although the latent code $\bm{z}$ is intended to capture subject-specific variations, the possibility remains for shared anatomical information to redundantly leak into $\bm{z}$. We address this empirically by tuning the dimensionality of $\bm{z}$, balancing reconstruction accuracy with the performance in downstream tasks (e.g., scan age prediction). As demonstrated in the ablation study (Table~\ref{tab:ablation_study}), a smaller latent dimension limits spatial representation but effectively captures subject-specific traits such as scan age, whereas a larger dimension enhances spatial accuracy for tasks like tissue segmentation.

\subsubsection*{Explicit vs. Learnable Conditioning}
Our experiments show that explicit conditioning on known clinical factors, such as birth age or LV volume, enhances interpretability by decoupling these factors from the latent code $\bm{z}$. However, so far this separation is not strictly enforced, theoretically allowing the network to ignore explicit conditions. While we have not encountered this phenomenon in our experiments, evaluating the network's ability to predict the conditioned variable during test-time adaptation, helps to gain insight on how well the network captures the conditioned variable. Furthermore, explicitly defined conditioned variables may limit flexibility and representation capacity. A promising extension is the adoption of a multitask learning framework in which condition representations are learned jointly with auxiliary tasks such as age prediction or abnormality classification. This would allow conditioning to emerge from the data itself, reducing dependence on predefined labels and improving generalizability. While such an approach increases model complexity and requires larger, well-annotated datasets, it represents a valuable direction for future work—particularly in large-scale population studies.

\subsubsection*{Spatial Latent Codes and Spatial Transformations}
The introduction of spatial latent codes, queried via trilinear interpolation, significantly enhances \cinema’s capability to represent complex anatomical features. However, experiments indicate limited gains from more sophisticated interpolation methods, such as convolutional neural networks (CNNs) in this particular setup. Future explorations into CNN or attention-based interpolation methods may reveal advantages in high-resolution or large-scale data contexts, such as adult cohorts. 
Spatial transformer networks (STNs) offer a promising approach to better model non-rigid spatial transformations during training and inference. While their integration introduces challenges—particularly in disentangling spatial transformations from latent codes—STNs could enhance alignment in anatomically diverse or highly abnormal subjects, making them a primising direction for future work.
Finally, the fixed Gaussian kernel for latent code regression during atlas inference may be suboptimal for highly unbalanced datasets with non-uniform distributions. Adopting adaptive kernel regression methods, as in \cite{Serag2012}, could further enhance atlas accuracy.

\subsubsection*{Topological Correctness}
\cinema~demonstrates superior accuracy in modeling brain tissue volume growth trajectories and cortical folding compared to atlases of established baselines (Fig.~\ref{fig:mm_neo_atlas}).  
Nonetheless, ensuring topological accuracy, particularly in cortical structures, remains challenging and \emph{cannot be guaranteed} in the current work. Future work should focus on integrating topology-preserving loss functions, e.g., via a patch-based regularization term, or incorporating cortical surface maps as additional input modality to explicitly enhance topological correctness.

\subsection{Future Applications}

\subsubsection*{Cross-Subject Normalization}
Cross-subject normalization—the alignment of anatomical data from different individuals into a common reference space for group-level analysis—is another application of atlas-based methods. Although not explicitly evaluated in this work, \cinema~supports this through conventional and hybrid strategies. A reference atlas can be generated from the cohort mean and used for registration. Additionally, for anatomically atypical subjects (e.g., severe ventriculomegaly), the subject's latent code can be iteratively interpolated toward the normative mean (similar to Fig.~\ref{fig:lat_code_ip}), with simultaneous flow-field estimation between interpolation steps. This enables smooth diffeomorphic deformation trajectories from abnormal to normal anatomies or vice versa.

Ultimately, \cinema~is designed to enable population analysis directly in latent space, where anatomical variability is compactly encoded and can be quantitatively assessed or mapped back to image space. This shifts the normalization paradigm from spatial deformation in image space to representational alignment in latent space, offering a flexible and powerful foundation for future atlas-based studies.

\subsubsection*{Extension to Additional Modalities}
Our framework naturally generalizes beyond T1- and T2-weighted MRI, accommodating various imaging modalities, including cortical surface maps or 4D data such as diffusion weighted imaging \cite{Consagra2024} or quantitative MRI. Handling these modalities merely involves additional input channels (e.g., diffusion direction or echo time) which enables joint modeling without structural changes to the framework. Optional domain-specific constraints (e.g., exponential decay for T2*) can be added to regularize learning. While computational cost increases, the extension is straightforward and presents a promising direction for future work.

\subsubsection*{Extended Latent Space Analysis}
Future work could also further explore \cinema’s latent space, which holds potential for broader applications, such as disease classification or anomaly detection.

\section{Conclusion}
\cinema~represents a significant advancement in spatio-temporal brain atlas generation, addressing limitations inherent in traditional and deep learning-based methods. By leveraging implicit neural representations, \cinema~achieves efficient, flexible, and high-resolution modeling of both typical and pathological neurodevelopmental patterns. Its compact latent space, explicit conditioning capabilities, and generative potential for synthetic data production significantly enhance its utility for neuroimaging research. Additionally, \cinema~serves as an out-of-the-box framework with minimal pre-processing demands, requiring neither prior atlases nor external registration procedures, constructing atlases in minutes rather than hours or days. While further refinements in topological accuracy and adaptive kernel strategies remain areas for future work, \cinema~already achieves higher anatomical fidelity and more accurate representations than established baselines, making it well suited for broad clinical and research applications.

\bibliographystyle{ieeetr}
\bibliography{CINeMA/references.bib}
\end{document}